\documentclass[conference]{IEEEtran}
\IEEEoverridecommandlockouts
\usepackage{cite}
\usepackage{amsmath,amssymb,amsfonts}
\usepackage{graphicx}
\usepackage{subcaption}
\usepackage{algorithm,algorithmic}
\usepackage{textcomp}
\usepackage{xcolor}
\usepackage{pifont}
\usepackage{multirow}
\usepackage{hyperref}
\newcommand{\circlednumber}[1]{\ding{\numexpr171+#1\relax}}
\usepackage{enumitem}
\usepackage{pifont}
\def\BibTeX{{\rm B\kern-.05em{\sc i\kern-.025em b}\kern-.08em
    T\kern-.1667em\lower.7ex\hbox{E}\kern-.125emX}}
\begin{document}

\title{OWLed: Outlier-weighed Layerwise Pruning for Efficient Autonomous Driving Framework 
\\
\thanks{Correspondence to Xilu Wang: wangxilu@surrey.ac.uk.}
}
\author{\IEEEauthorblockN{Jiaxi Li}
\IEEEauthorblockA{\textit{Computer Science Research Centre} \\
\textit{University of Surrey}\\
Guildford, United Kingdom 
}
\and
\IEEEauthorblockN{Lu Yin}
\IEEEauthorblockA{\textit{Computer Science Research Centre} \\
\textit{University of Surrey}\\
Guildford, United Kingdom 
}
\and
\IEEEauthorblockN{Xilu Wang}
\IEEEauthorblockA{\textit{Computer Science Research Centre} \\
\textit{University of Surrey}\\
Guildford, United Kingdom 
}
}

\maketitle
\begin{abstract}
The integration of Large Language Models (LLMs) into autonomous driving systems offers promising enhancements in environmental understanding and decision-making. However, the substantial computational demands of deploying LLMs locally on vehicles render this approach unfeasible for real-world automotive applications. To address this challenge, we introduce OWLed, the \underline{O}utlier-\underline{w}eighed \underline{L}ayerwise Pruning for
\underline{E}fficient Autonomous \underline{D}riving Framework that leverages outlier-weighed layerwise sparsity for model compression. Our method assigns non-uniform sparsity ratios to different layers based on the distribution of outlier features, significantly reducing the model size without the need for fine-tuning. To ensure the compressed model adapts well to autonomous driving tasks, we incorporate driving environment data into both the calibration and pruning processes. Our empirical studies reveal that the encoder component is more sensitive to pruning than the LLM, highlighting its critical role in the system. Experimental results demonstrate that OWLed outperforms existing methods in perception, action prediction, and language understanding while substantially lowering computational requirements. These findings underscore the potential of combining advanced pruning techniques with LLMs to develop efficient and robust autonomous driving systems capable of handling complex scenarios. 
Code is available at \url{https://github.com/JiaxiLi1/OWLed}.
\end{abstract}


\section{Introduction}
\label{sec: introduction}
Autonomous driving technology has made significant progress in recent years, with an increasing number of autonomous vehicles being tested and deployed on public roads~\cite{teng2023motion, chen2024end}. There are generally two types of autonomous driving systems: one is a modular design with different sub-modules to complete various tasks such as perception, prediction, and planning~\cite{casas2020implicit, li2022bevformer, shi2022motion, zhou2024smartrefine}. While this modular approach provides system interpretability, it suffers from error accumulation and information loss between components because its modules cannot directly access sensor data. The other is an end-to-end design that directly inputs sensor data into a series of neural networks to obtain control signals~\cite{hu2023planning, shao2023reasonnet, chen2024driving}. However, systems built with traditional neural networks often struggle to handle long-tail scenarios and complex urban environments, limiting their ability to process the diverse and unpredictable situations encountered in real-world driving~\cite{teng2023motion, chen2024end, cui2024survey}. 

Recently, large language models (LLMs) have exhibited capabilities approaching Artificial General Intelligence, including common sense understanding, reasoning, and knowledge retrieval~\cite{NEURIPS2020_1457c0d6, bubeck2023sparks, touvron2023llama}. LLMs' remarkable capabilities make them a promising solution for addressing the aforementioned challenges in autonomous driving systems. This has sparked several attempts to integrate LLMs into autonomous vehicles to enhance understanding of the driving environment, explain perception results \cite{sha2023languagempc}, generate textual driving decisions \cite{xu2024drivegpt4}, and translate those decisions into executable commands~\cite{cui2024survey}. These works primarily fall into two categories: The first category leverages pre-trained LLMs, such as ChatGPT, through API interfaces to process sensor data (i.e., camera and radar inputs) for environmental understanding and decision-making~\cite{sha2023languagempc, mao2023gpt, wen2024road, xu2024drivegpt4, wen2024dilu}. 
However, its drawback lies in its dependence on network conditions and the data processing speed of servers, which limits its effectiveness in complex driving scenarios~\cite{cui2024survey}. The second category involves training and deploying LLMs locally on autonomous vehicle systems. This method could address the autonomous driving system's requirements for cloud computing resources and network communication quality~\cite{nie2023reason2drive, ma2023dolphins, chen2024driving, tian2024drivevlm, shao2024lmdrive, guo2024co, ding2024holistic}. However, as LLMs typically operate at the billion-parameter scale, this approach places substantial demands on local computational resources, thereby limiting their practical applications in autonomous vehicles.

To address the computational challenges posed by deploying LLMs locally, numerous studies have explored various model compression techniques for applications with limited computational budget, including quantization \cite{kim2024memory}, knowledge distillation \cite{gu2024minillm} and network pruning \cite{liu2018rethinking}. Among them, network pruning has emerged as a promising approach to improve the efficiency of deep networks by reducing model size~\cite{mozer1988skeletonization, janowsky1989pruning, le1990optimal, han2015learning, han2015deep}. Specifically, pruning methods aim to reduce the model size by removing unnecessary weights and maintaining performance. However, its application to LLMs is limited because traditional pruning methods often require fine-tuning or retraining to obtain the optimal performance, which is prohibitively expensive for LLMs due to their massive size and requirements for sheer amount of computational resources~\cite{liu2018rethinking,gale2019state,renda2020comparing,blalock2020state}. 

Recent advancements have explored pruning LLMs without fine-tuning. These studies have demonstrated that a substantial portion of the model's parameters can be eliminated in a single step while incurring only minimal performance degradation. For example, sparseGPT~\cite{frantar2023sparsegpt} is a one-shot pruning method based on layer-wise reconstruction. It uses the Optimal Brain Surgeon (OBS) algorithm to determine which weights to be pruned. Wanda~\cite{sun2023a}, on the other hand, is a simple yet effective pruning approach that considers both weight magnitudes and input activations. Wanda is computationally more efficient than SparseGPT as it doesn't require the approximation of second-order information. Note that both SparseGPT and Wanda are designed to apply uniform layerwise sparsity when pruning LLMs. This means that each layer in the model is pruned to the same degree of sparsity, regardless of its position or function within the network. More recently, research has shown that non-uniform layerwise sparsity can potentially yield better performance~\cite{pmlr-v235-yin24e}. Building on this insight, Yin \emph{et al.}~\cite{pmlr-v235-yin24e} proposed OWL as a non-uniform layerwise sparsity strategy based on the distribution of outlier features. OWL captures the impact of outlier features in LLMs and assigns sparsity ratios proportional to each layer's outlier ratio. OWL has shown demonstrated superior performance compared to uniform pruning methods, particularly at high sparsity levels. Importantly, it can be combined with existing pruning techniques to further improve performance.

While many attempts have been made to leverage LLMs in autonomous driving, the research of computationally efficient LLMs for end-to-end autonomous driving is still at the very preliminary stage. To address the computational challenges associated with local LLM deployment in autonomous vehicles, we propose the first effective LLM-assisted autonomous driving framework by utilizing outlier-weighed layerwise sparsity to compress LLMs, termed OWLed. To ensure effective adaptation of LLMs to the specific demands of autonomous driving, we incorporate downstream task data into both the calibration process and the optimization of the pruning procedure. This tailored approach allows for more precise and context-specific model compression. The main contributions of this paper are summarized as follows:

\begin{enumerate}[label=\circlednumber{\arabic*},labelindent=\parindent]
\item We propose an effective LLM assisted autonomous driving framework that significantly enhances computational efficiency while maintaining the powerful reasoning capabilities of LLMs. By employing outlier-weighed layerwise sparsity, our approach effectively addresses the computational constraints inherent in deploying LLMs in vehicles. This framework is designed to process and reason over complex environmental data from multiple sensory modalities and natural language inputs, enabling more sophisticated and efficient decision-making in autonomous vehicles.
\item To further enhance the performance of OWLed, we introduce a novel calibration strategy tailored specifically for autonomous driving applications. While current pruning methods rely solely on generic text data for calibration, our approach leverages diverse data collected from real driving environments. This novel use of domain-specific calibration data allows us to better adapt the LLM to autonomous driving.
\item We investigated the importance of the encoder and the LLM in the context of pruning for autonomous driving systems. Our empirical results show that the encoder plays a crucial role in these systems and exhibits higher sensitivity to pruning compared to the LLM component. 

\item Compared to existing methods, our framework demonstrates superior perception, action prediction, and language understanding capabilities while significantly reducing computational requirements. We validate the effectiveness of our method through extensive experiments and ablation studies.
\end{enumerate}

The rest of this paper is structured as follows: Section~\ref{sec: related work} presents the related work for LLMs-based autonomous driving and pruning methods. The proposed method is detailed in Section~\ref{sec: method}. Section~\ref{sec: experiments} presents the main experimental setup, results, and the ablation study. We conclude the paper in Section~\ref{sec: conclusion}.

\section{Related work}
\label{sec: related work}
\subsection{LLMs-based autonomous driving systems}
\label{subsec: LLMs based autonomous driving systems}
The integration of LLMs into autonomous driving systems is a new approach to improving the reasoning ability, interpretability, and decision-making performance of the system. These works can be mainly divided into two categories: utilizing pre-trained LLMs via API interfaces, and training and deploying LLMs locally on vehicles.

API-based methods have been demonstrated flexibility in leveraging pre-trained LLMs for diverse driving tasks. For instance, LanguageMPC~\cite{sha2023languagempc} utilizes LLMs to generate high-level driving strategies by reasoning over complex scenarios, which are then translated into low-level control signals through a parameter matrix. This approach showcases the potential of LLMs in bridging the gap between abstract reasoning and concrete vehicle control. Similarly, GPT-Driver~\cite{mao2023gpt} reformulates motion planning as a natural language modeling task by converting heterogeneous scene input into language tokens. DRIVEGPT4~\cite{xu2024drivegpt4} proposes a multi-modal LLM framework that receives a series of image frames as input and then generates responses to human inquiries and predicts control signals for the next step.

Alternatively, local deployment approaches, while computationally demanding, offer advantages in terms of latency and data privacy. LLM-Driver~\cite{chen2024driving} integrating vectorized object-level 2D scene representations with locally deployed LLMs. This method enables real-time question answering about the driving environment and enhances the system's ability to explain its decisions to users. LMDrive~\cite{shao2024lmdrive} uses a vision encoder to process images and generate visual tokens. It is an end-to-end, closed-loop autonomous driving framework that is designed to process multi-modal sensor data (including cameras) with natural language instructions.

However, these methods face limitations due to their reliance on cloud server connections or powerful local computational resources for LLM deployment. Additionally, both approaches result in extended computation times, hindering real-time data processing. To address this issue, we pioneer the application of pruning methods to reduce the size of LLMs' parameters for autonomous driving tasks, enabling deployment on more onboard vehicle systems.
\subsection{Neural network pruning}
\label{subsec: pruning for LLMs}
Deep neural network pruning has proven effective in enabling high-performance perception models to be deployed on autonomous driving platforms with limited computational and memory resources. However, most existing pruning works focus on CNN-based autonomous driving systems \cite{zhao2021neural, lu2024crossprune, vemparala2020l2pf}. The integration of LLMs with autonomous driving systems, especially in the context of pruning for efficiency, remains largely unexplored.

 Traditional LLM pruning methods usually require fine-tuning or retraining to maintain performance. For example, LLM-Pruner is proposed in~\cite{ma2023llm} for structural pruning. It uses Taylor pruning to remove entire weight rows based on their impact on the model output.  Recent research has pivoted towards developing LLM pruning methods that do not require fine-tuning. SparseGPT~\cite{frantar2023sparsegpt} uses the OBS algorithm to select weights for pruning and then updates the remaining weights to minimize output reconstruction error. In this way, it addresses the challenge of computing the Hessian inverse. Wanda~\cite{sun2023a} evaluates an importance score for each weight by calculating the product of weight magnitude and the $\ell$2 norm of associated input features. Accordingly, weights can be pruned based on the importance score, without relying on second-order information or weight updates. Dynamic sparsity~\cite{evci2020rigging, liu2021we} is extended in~\cite{zhang2024dynamic} to efficiently fine-tune sparse LLM without weight updating, while the JunK DNA hypothesis \cite{yinjunk} investigates optimal overall sparsity for different downstream tasks. However, most existing methods apply uniform sparsity across all layers, overlooking the presence of outlier features in LLMs. Inspired by this, OWL~\cite{pmlr-v235-yin24e} introduces a non-uniform layerwise sparsity strategy. This approach calculates each layer's Layerwise Outlier Distribution (LOD) and assigns sparsity ratios proportional to these outliers, demonstrating promising performance gain.

Inspired by OWL, we leverage outlier information to compute optimal sparsity within an autonomous driving framework. Our method utilizes the insight that different layers contribute differently to the model's decision-making process in autonomous driving scenarios. By identifying and preserving crucial outlier features specific to driving tasks, OWLed maintains high performance even at high sparsity levels. To the best of our knowledge, this is the first work to integrate LLMs with autonomous driving systems in the context of pruning for efficiency.


\section{OWLed: Outlier-weighed Layerwise Pruning for Efficient Autonomous Driving Framework}
\label{sec: method}
\begin{figure}[htbp]
	\centering
        \hspace{0.55cm} 
	\begin{subfigure}[t]{0.9\linewidth}
		\includegraphics[width=\textwidth]{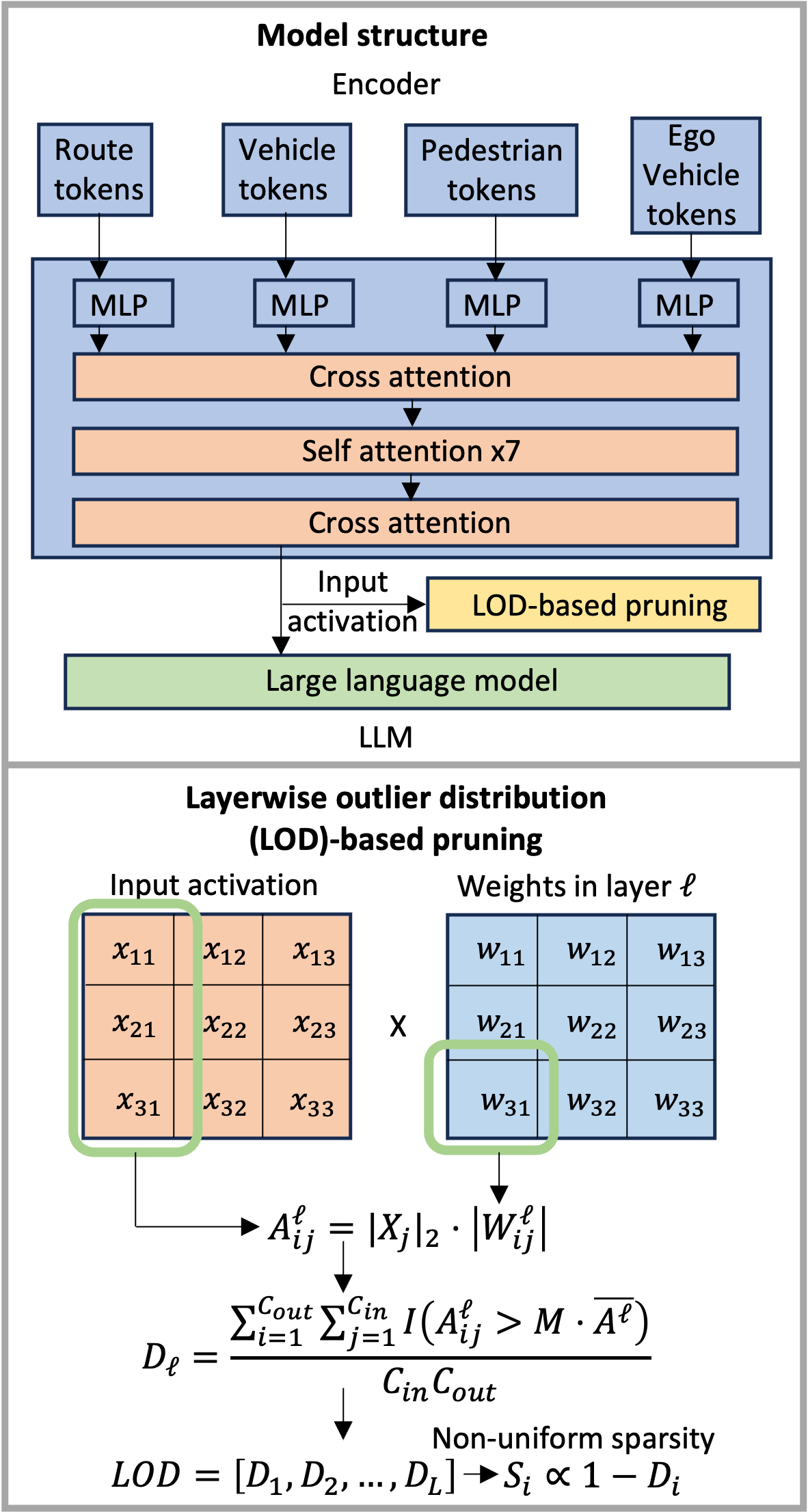}
	\end{subfigure}\par
	\caption{Illustration of 
 the architecture of OWLed and the LOD-based pruning. The autonomous driving model consists of a vector encoder for processing multimodal input data and an LLM for decision making. The model uses outlier-weighed layer-wise pruning and incorporates input activation to achieve efficient compression while maintaining the model performance.}
	\label{fig: method}
\end{figure}

In this section, we introduce OWLed, an efficient autonomous driving framework. Figure \ref{fig: method} illustrates the framework of OWLed. The corresponding pseudocode is provided in Algorithm \ref{alg:owled}.
\begin{algorithm}[htbp]
	\begin{algorithmic}[1]
		\caption{OWLed}
		\label{alg:owled}
            
            \vspace{0.05em}
            \STATE \textbf{Input:}
					$A$: Calibration data,
					$M$: Threshold coefficient,
					$\lambda$: Limiting value,
                    $L$: Number of layers;
            \vspace{0.1em}
            \STATE \textbf{Pruning}
            \FOR{$i=1$ to $L$}
		\STATE Use Equation \ref{eq1} to calculate outlier ratio $D_\ell$: \\
                    $D_\ell = \frac{\sum_{i=1}^{C_{out}} \sum_{j=1}^{C_{in}} I(A^\ell_{ij} > M \cdot \bar{A^\ell})}{C_{in}C_{out}}$;
		\ENDFOR
            \STATE Calculate target sparsity $[S_1, S_2, ..., S_{L}]$ by \\ using $S_i \propto 1 - D_i$;
            \STATE Applying pruning method;
            \STATE \textbf{Inference}
            \STATE Input vector token to encoder to obtain vector embedding;
            \STATE Combine vector embedding and prompt embedding;
            \STATE Input collected embedding to LLM.
  \vspace{0.5em}
  \STATE \textbf{Output:} LLM responses.
	\end{algorithmic}
\end{algorithm}
\subsection{LLMs-based autonomous driving framework}
\label{subsec: LLMs-based autonomous driving framework}
OWLed adopts the architecture proposed in \cite{chen2024driving}. This architecture consists of two key components: a vector encoder and an LLM. The vector encoder first processes multi-modal input data, including environment information about routes, vehicles, pedestrians, and the ego vehicle, through Multilayer Perceptron (MLP) layers. These processed vectors are then projected into a latent space via a cross-attention layer, with ego features added to each learned input latent vector to emphasize the ego state. Subsequently, a self-attention module followed by a cross-attention layer transforms these encoded representations into embeddings compatible with the LLM. After obtaining the encoded vector representations, we insert them into a text prompt embedding to create the input for the LLM. Finally, the LLM generates driving decisions and answers queries.

In our experiments, LLaMA-7b \cite{touvron2023llama} with pre-trained model weights is adopted as the LLM. To focus on the impact of pruning methods on model performance, we removed the Low-Rank Adaptation (LoRA) module used in \cite{chen2024driving} and instead merged its fine-tuned weights with LLaMA-7b.

\subsection{Layerwise Outlier Distribution-based pruning}
\label{subsec: Outlier Weighed Layerwise Sparsity}

To effectively prune the LLM while maintaining model performance, we adopt the Layerwise Outlier Distribution (LOD) proposed in \cite{pmlr-v235-yin24e} to quantify the outlier distribution across layers. This method quantifies the outlier distribution across layers, allowing us to align each layer's sparsity ratio with its outlier ratio. By doing so, we aim to retain key outlier features crucial for maintaining model performance.

We first calculate the LOD of each layer of the model. For a model with $L$ layers, we calculate LOD = $[D_1, D_2, ..., D_{L}]$. Specifically, for the $\ell$th layer, we calculate its outlier ratio $D_\ell$ as follows:

\begin{equation}\label{eq1}
D_\ell = \frac{\sum_{i=1}^{C_{out}} \sum_{j=1}^{C_{in}} I(A^\ell_{ij} > M \cdot \bar{A^\ell})}{C_{in}C_{out}}
\end{equation}
where $A^\ell_{ij} = \|X_j\|_2 \cdot |W^\ell_{ij}|$ is the outlier score of the weight $W^\ell_{ij}$, calculated as the $\ell_2$ norm of all input features connected to this weight multiplied by the absolute value of the weight. $\bar{A^\ell}$ is the average of all outlier scores of the layer, $M$ is a hyperparameter used to determine the threshold of the outlier, which is determined through grid search on the validation set (see Section~\ref{sec: experiments} for details). $I(\cdot)$ denotes the indicator function, which returns 1 when the condition is met and 0 otherwise. $C_{in}$ and $C_{out}$ denote the input and output channel dimensions of the layer, respectively.

Given the target model sparsity $S$, we determine the sparsity $[S_1, S_2, ..., S_{L}]$ of each layer, based on the principle that layers with higher outlier ratios should have lower sparsity. Specifically, we set $S_i \propto 1 - D_i$. To prevent the sparsity difference between layers from being too large, we introduce a hyperparameter $\lambda$ to limit the sparsity of each layer to the range of $[S - \lambda, S + \lambda]$, while keeping the average sparsity of all layers as $S$.

\subsection{Calibration data for OWLed}
Traditional pruning methods~\cite{pmlr-v235-yin24e, sun2023a,frantar2023sparsegpt} utilize random text snippets from the C4 datasets, which contains general linguistic patterns but lacks driving-specific semantics. 

Regarding the calculation of LOD, in\cite{pmlr-v235-yin24e}, the LOD is calculated using data from the C4 dataset \cite{raffel2020exploring} as calibration data. However, in our autonomous driving context, the input data consists of vector representations of driving scenarios, which significantly differs from general text data. This disparity in input data can lead to substantially different LOD calculations, potentially affecting the pruning results. To better align with our downstream task requirements and to leverage task-specific information, we propose to use our driving scenario vector data to calculate the LOD values for pruning. This approach ensures that the pruning process is tailored to the specific needs and characteristics of our autonomous driving task, potentially leading to more effective pruning results.

\begin{figure*}[htbp]
	\centering
	\begin{subfigure}[t]{\linewidth}
		\includegraphics[width=\textwidth]{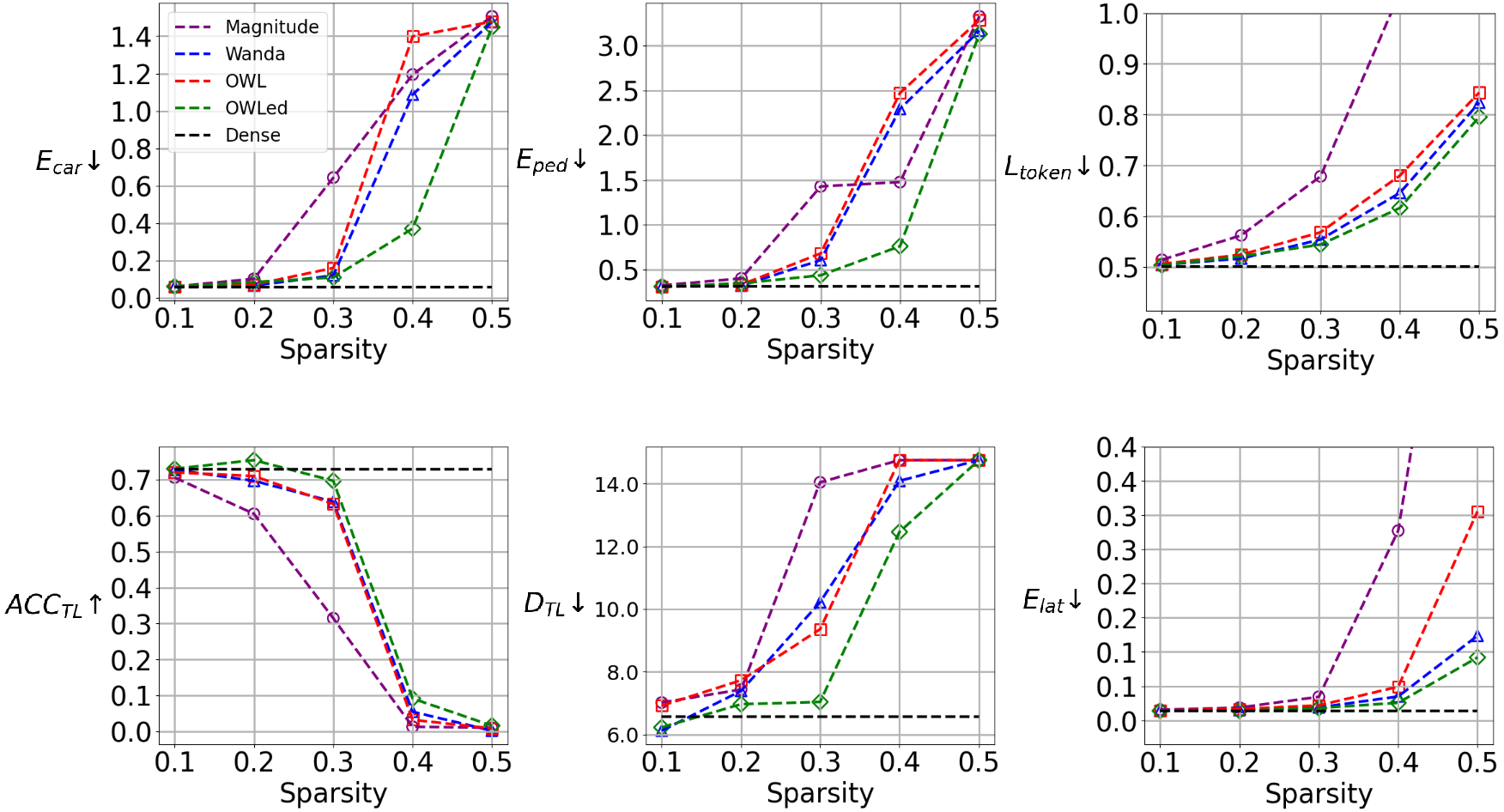}
		\label{fig: xxx_a}
	\end{subfigure}

	\caption{The evaluation results of perception and action prediction achieved by different pruning techniques. Results show performance across multiple metrics including car detection error ($E_{car}$), pedestrian detection error ($E_{ped}$), token prediction loss ($L_{token}$), traffic light detection accuracy ($ACC_{TL}$), traffic light distance error ($D_{TL}$), and lateral control error ($E_{lat}$) at sparsity ratios varying from 0.1 to 0.5.}
	\label{fig: perception and action prediction}
\end{figure*}
	
	

\section{Experiments}
\label{sec: experiments}

\subsection{Experiment setup}
\label{subsubsec: experiment setup}
In this section, we verify the performance of OWLed. We explore two types of calibration data for calculating LOD. Traditional pruning methods typically use general text data from the C4 dataset~\cite{raffel2020exploring}, which consists of a diverse range of web-crawled text. In contrast, we propose using domain-specific driving scenario data collected from a custom-built 2D simulator~\cite{chen2024driving}. This dataset includes continuous driving data from 15 diverse virtual environments with randomly generated traffic conditions, containing vector representations of driving scenarios that are directly relevant to our autonomous driving tasks. The use of task-specific calibration data allows for more targeted and effective pruning in the context of autonomous driving.

We employ the driving QA dataset in~\cite{chen2024driving} as our evaluation dataset. The dataset comprises three subsets: a 100k dataset for vector encoder pretraining, a 10k dataset for fine-tuning, and a 1k dataset for evaluation.

We adopt the evaluation metrics in~\cite{chen2024driving} to evaluate the proposed model's perception and action prediction ability.
These metrics include mean absolute error for the number of cars prediction ($E_{car}$) and pedestrians prediction ($E_{ped}$), accuracy of traffic light detection ($ACC_{TL}$), mean absolute distance error in meters for traffic light distance prediction ($D_{TL}$) and normalized steering wheel angle ($E_{lat}$), and weighted cross-entropy loss for token prediction on the evaluation set ($L_{token}$). Additionally, we utilize GPT-3.5 to evaluate the quality of the model's responses to the driving environments. This technique allows for rapid and consistent assessment of answers without fixed solutions~\cite{fu2023gptscore, wang2023chatgpt, liu2023g}. Specifically, we provide GPT-3.5 with the language-based observation description used during dataset generation, test questions, and the model's answer. Based on the given observations, the GPT then generates an assessment and a score from 0 to 10 for each response. The results of the GPT evaluations are the model's final score averaged across all questions.
\begin{table}[ht!]
\centering
\begin{tabular}{ccc}
\hline\hline
Sparsity ratio & Method & GPT Grading $\uparrow$ \\
\hline
\multirow{4}{*}{0.3} & Magnitude & 7.03 \\
& Wanda & 7.24 \\
& OWL & 7.32 \\
& OWLed & \textbf{7.53} \\
\hline
\multirow{4}{*}{0.4} & Magnitude & 4.68 \\
& Wanda & 6.14 \\
& OWL & 6.36 \\
& OWLed & \textbf{6.94} \\
\hline\hline
& Dense & 8.45 \\
& Constant answer "I don't know" & 2.92 \\
& Randomly shuffled answers & 3.88 \\
& Answers generated by GPT & 9.47 \\
\hline\hline
\end{tabular}
\caption{GPT grading results of the driving QA outputs achieved by different pruning methods at sparsity ratios of 0.3 and 0.4, and baseline performances achieved by the dense model, constant answers, random answers, and GPT-generated answers. Higher scores indicate better quality of responses to driving-related questions.}
\label{tab: gpt-grading1}
\end{table}

\begin{table*}[!]
\centering
\begin{tabular}{ccccccccc}
\hline\hline
Sparsity ratio & Method & $E_{car}\downarrow$ & $E_{ped}\downarrow$ & $L_{token}\downarrow$ & $ACC_{TL}\uparrow$ & $D_{TL}\downarrow$ & $E_{lat}\downarrow$ \\
\hline
\multirow{3}{*}{0.3} & OWLed & \textbf{0.109} & \textbf{0.434} & \textbf{0.544} & 0.697 & \textbf{7.040} & \textbf{0.018} \\
& OWLed-separate & 0.275 & 0.733 & 0.587 & 0.639 & 8.444 & 0.020 \\
& OWLed-global & 0.221 & 0.630 & 0.603 & \textbf{0.706} & 7.508 & 0.022 \\
\hline
\multirow{3}{*}{0.4} & OWLed & \textbf{0.371} & \textbf{0.760} & \textbf{0.616} & 0.092 & 12.461 & \textbf{0.026} \\
& OWLed-separate & 0.483 & 0.997 & 0.722 & \textbf{0.187} & \textbf{9.879} & 0.040 \\
& OWLed-global & 0.479 & 0.999 & 0.718 & 0.088 & 14.100 & 0.042 \\
\hline\hline
\end{tabular}
\caption{Evaluation of pruning encoder ablation. Comparison between OWLed variants (OWLed, OWLed-separate, OWLed-global) across different metrics at sparsity ratios of 0.3 and 0.4, respectively.}
\label{tab: ablation study1}
\end{table*}

\subsection{Main experiments}
\label{subsec: main experiments}
\subsubsection{Baselines}
\label{subsubsec: baselines}
The original autonomous driving model from~\cite{chen2024driving} is used as one of our baselines, referred to as Dense. Additionally, we create three more baselines by adopting different state-of-the-art pruning techniques to this architecture, i.e., Magnitude~\cite{jaiswal2024emergence}, Wanda~\cite{sun2023a} and OWL~\cite{pmlr-v235-yin24e}. We test all the methods under comparison with different sparsity levels, i.e., $ sparsity ratio \in (10\%,20\%,30\%,40\%,50\%)$, and set the number of calibration data as 128 for all baselines. For the OWL implementation, we conduct a comprehensive grid search to optimize the hyperparameters, selecting $\lambda=0.1$ and $M=5$ from candidate ranges of $\lambda \in(\text{0.02,0.05,0.08,0.1,0.2})$ and $M \in(\text{3, 5, 7, 10})$, following \cite{pmlr-v235-yin24e}.
\subsubsection{Results}
\label{subsubsec: Results}
The experimental results for perception and action prediction, measured using various evaluation metrics across different levels of layerwise sparsity, are shown in Figure \ref{fig: perception and action prediction}. The results of GPT grading responses achieved by all algorithms under comparison are summarized in Table \ref{tab: gpt-grading1}. 

Figure \ref{fig: perception and action prediction} demonstrates that OWLed consistently outperforms other methods across all evaluation metrics. At sparsity levels of 0.1 and 0.2, all models show only a slight performance decrease compared to the Dense model. The Magnitude method, however, exhibits an expected performance decline as sparsity increases, with a notable drop at 0.3 sparsity. This decline stems from its reliance solely on weight magnitude for pruning, neglecting other crucial aspects and failing to capture outlier values. At sparsity levels above 0.3, the Magnitude method's performance deteriorates dramatically across multiple evaluation metrics, most notably in $L_{token}$ and $E_{lat}$. 

Wanda and OWL demonstrate similar performance. Both methods consider input activation when calculating weight outlier values for pruning. Additionally, OWL employs non-uniform sparsity pruning by calculating the LOD. However, OWL's performance could be influenced by its hyperparameter, which leads to its effectiveness can not be guaranteed across different tasks. For example, while OWL achieves comparable performance in terms of $ACC_{TL}$, it fails to maintain its efficiency for the steering wheel angle prediction, as indicated by $E_{lat}$. We note that for the $ACC_{TL}$ metric, OWLed even outperforms the dense model at 0.2 sparsity, which suggests that the dense model is not always better than the sparse model. This could be because of the sparse model's ability to prevent overfitting by decreasing the capacity of the model~\cite{ahmad2019can, zhang2021can}.

OWLed consistently outperforms other methods under comparison across all evaluation metrics, validating the effectiveness of the use of LOD and its calibration strategy specifically designed for autonomous driving scenarios. Table \ref{tab: gpt-grading1} presents GPT grading results that further confirm OWLed's advantage in autonomous driving applications. While the performance gaps between models are relatively small at sparsity levels of 0.1, 0.2, and 0.5, we focus on GPT evaluations of model responses at 0.3 and 0.4 sparsity for a more detailed analysis. At these levels, OWLed achieves higher scores than other methods, with its performance gain increasing at higher sparsity levels. Moreover, OWLed's scores are superior to incorrect answers ("I don't know" and randomly generated responses) but lower than those of the Dense model and the GPT-provided answers, which serve as performance baselines. These results further validate the effectiveness of our proposed model.

\subsection{Ablation study: impact of pruning model encoder}
\label{subsec: ablation study1}
Our model consists of two main components: an encoder and LLaMA-7B as the LLM. The encoder has 9.1e+6 parameters, while LLaMA-7B has 6.7e+9 parameters. The encoder's parameter size is approximately three orders of magnitude smaller than the LLM's. Given this significant size difference, we conduct an ablation study to investigate how pruning the relatively compact encoder affects overall model performance.

In addition to OWLed, which applies LOD-based pruning only to LLaMA, we introduce two baselines: OWLed-separate and OWLed-global. OWLed-separate calculates LOD and applies pruning separately for the encoder and LLM. OWLed-global, on the other hand, calculates outlier values across all layers before determining the target sparsity, followed by LOD-based pruning. 
\vspace{0.5em}
\begin{table}[ht]
\centering
\begin{tabular}{ccc}
\hline\hline
Sparsity ratio & Method & GPT Grading $\uparrow$ \\
\hline
\multirow{3}{*}{0.3} & OWLed & \textbf{7.53} \\
& OWLed-separate & 7.23 \\
& OWLed-global & 7.29 \\
\hline
\multirow{3}{*}{0.4} & OWLed & \textbf{6.94} \\
& OWLed-separate & 6.47 \\
& OWLed-global & 6.51 \\
\hline\hline
\end{tabular}
\caption{GPT grading results of language understanding capabilities across OWLed variants at 0.3 and 0.4 sparsity ratios, respectively.}
\label{tab:ablation study1 gpt grading}
\end{table}

\begin{table}[htbp]
\centering
\small
\begin{tabular}{@{}c@{\hspace{2mm}}c@{\hspace{2mm}}c@{\hspace{2mm}}c@{\hspace{2mm}}c@{\hspace{2mm}}c@{\hspace{2mm}}c@{}}
\hline\hline
Calibration \\ Samples & $E_{car}$ & $E_{ped}$ & $L_{tok}$ & $ACC_{TL}$ & $D_{TL}$ & $E_{lat}$ \\
 & $\downarrow$ & $\downarrow$ & $\downarrow$ & $\uparrow$ & $\downarrow$ & $\downarrow$ \\
\hline
32 & 0.554 & 1.592 & 0.702 & 0.071 & 15.441 & 0.037 \\
64 & 0.327 & \textbf{0.693} & 0.701 & 0.061 & 14.963 & 0.030 \\
128 & 0.371 & 0.759 & 0.699 & 0.091 & \textbf{14.100} & \textbf{0.025} \\
256 & \textbf{0.302} & 0.703 & 0.693 & 0.167 & 14.586 & 0.026 \\
512 & 0.342 & 0.726 & \textbf{0.689} & \textbf{0.254} & 14.572 & 0.030 \\
\hline\hline
\end{tabular}
\caption{Evaluation for calibration samples ablation. Performance comparison across different numbers of calibration samples (32 to 512) on various metrics.}
\label{tab:ablation study2 calibration samples evaluation}
\end{table}
\vspace{0.5em}
To ensure a fair comparison of different pruning methods, we establish a consistent baseline for parameter reduction. To achieve this, we first calculate the number of parameters pruned from LLaMA by OWLed at sparsity levels of 0.3 and 0.4 and then use this value as a coefficient. When pruning the encoder using OWLed-separate and OWLed-global, we scale the pruned parameters by the coefficient to ensure that OWLed, OWLed-separate, and OWLed-global prune the same number of parameters at the same sparsity level. This approach allows us to compare the performance of different methods under equivalent conditions. We report the results in Tables \ref{tab: ablation study1} - \ref{tab:ablation study1 gpt grading}.

As shown in Table \ref{tab: ablation study1}, at sparsity levels of 0.3 and 0.4, OWLed outperforms both OWLed-separate and OWLed-global across multiple evaluation metrics. The results in Table \ref{tab:ablation study1 gpt grading} are consistent with the trends observed in Table \ref{tab: ablation study1}. This indicates that pruning the encoder has a significant impact on model performance when pruning the same number of parameters. 

A possible explanation for the encoder's sensitivity to pruning is because of its role in translating sensor data into compact vector representations. Even minor distortions in these representations can propagate through the LLM, significantly impacting decision-making. In contrast, the LLM primarily processes high-level features and is acceptable to reduce part of the parameter due to its massive parameter count (6.7B vs. encoder's 9.1M parameters).

Therefore, in this scenario, we do not recommend pruning the encoder. Notably, since the encoder's parameter number is much less than that of the LLM, focusing on pruning the LLM may lead to better model performance.

\subsection{Ablation study: number of calibration samples}
\label{subsec: ablation study2}
The calibration strategy consists of calibration data type and calibration data number. An ablation study is conducted to investigate the impact of the number of calibration data on the OWLed performance. We vary the number of calibration data from the set $(\text{32, 64, 128, 256, 512})$ and present the results in Tables \ref{tab:ablation study2 calibration samples evaluation} and \ref{tab:calibration samples gpt}. We set the sparsity ratio as 0.4.

As shown in Table \ref{tab:ablation study2 calibration samples evaluation}, using only 32 calibration samples results in suboptimal performance across all metrics, this could because insufficient samples fail to capture the full distribution of feature importance across different driving scenarios. Although increasing samples beyond 32 improves performance, the gains become marginal after 128 samples. We recommend using 256 calibration samples in this task as an optimal balance between performance and computational efficiency.

Table \ref{tab:calibration samples gpt} shows that while 32 calibration samples yield the lowest GPT grading score, the scores for larger sample sizes are comparatively similar. This aligns with our observations from Table \ref{tab:ablation study2 calibration samples evaluation}: Beyond the threshold of 128 samples, increasing the number of calibration samples does not lead to a significant improvement in model performance. The consistency between the two evaluation methods strengthens our recommendation to select a moderate number of calibration samples, rather than simply increasing the number of samples. This approach effectively balances model performance with computational efficiency, optimizing the calibration process without incurring unnecessary computational overhead.

\vspace{0.5em}

\begin{table}[!]
\centering
\begin{tabular}{cc}
\hline\hline
Calibration samples & GPT Grading $\uparrow$ \\
\hline
32 & 5.63 \\
64 & 6.67 \\
128 & 6.94 \\
256 & \textbf{7.01} \\
512 & 6.88 \\
\hline\hline
\end{tabular}
\caption{GPT grading results of calibration samples ablation. Analysis of model response quality across different calibration sample sizes (32 to 512).}
\label{tab:calibration samples gpt}
\end{table}
\vspace{0.5em}

\section{Conclusion}

This paper presents OWLed, an efficient autonomous driving framework that employs outlier-weighed layerwise pruning to compress LLM without fine-tuning. By assigning non-uniform sparsity ratios based on the distribution of outlier features and incorporating driving environment data into calibration and pruning, our method significantly reduces computational demands while maintaining performance. We also found that the encoder is more sensitive to pruning than the LLM. Our experiments show that OWLed outperforms existing methods in perception, action prediction, and language understanding. These findings demonstrate the potential of combining advanced pruning techniques with LLMs to develop efficient and robust autonomous driving systems capable of handling complex scenarios. The OWLed's effectiveness may be constrained in scenarios with highly dynamic environments or unseen driving conditions, where the pre-computed outlier distributions might not fully capture the complexity of the situation. Future work includes applying weight-low-rank and quantization techniques to further increase model efficiency.

            

\label{sec: conclusion}

\vspace{12pt}

\bibliographystyle{IEEEtran}
\bibliography{ref} 

\end{document}